\documentclass[12pt, a4paper]{article}
\usepackage[font=footnotesize]{caption}
\usepackage[utf8]{inputenc}
\usepackage{graphicx}
\usepackage{authblk}
\usepackage{booktabs}
\usepackage{verbatim}
\usepackage{url}
\usepackage{caption}
\usepackage{subcaption}
\usepackage{indentfirst}
\usepackage{url}
\usepackage{amssymb}
\usepackage{amsmath}
\usepackage{array}
\usepackage{tcolorbox}
\title{Supervised Pattern Recognition Involving Skewed Feature Densities}

\author{Alexandre Benatti$^1$ and Luciano da F. Costa$^2$}

\affil{
$^1$Institute of Mathematics and Statistics - DCC \\
University of S\~ao Paulo \\
Rua do Mat\~ao, 1010, \\ S\~ao Paulo, SP 05508-090 Brazil 
\\ \vspace{0.5cm}
$^2$S\~ao Carlos Institute of Physics - DFCM \\
University of S\~ao Paulo \\
Av.~Trabalhador S\~ao-carlense, 400, \\ S\~ao Carlos, SP 13566-590 Brazil
}

\date{\emph{10th Aug., 2024}}

\begin{document}

\maketitle

\begin{abstract}
Pattern recognition constitutes a particularly important task underlying a great deal of scientific and technologica activities. At the same time, pattern recognition involves several challenges, including the choice of features to represent the data elements, as well as possible respective transformations. In the present work, the classification potential of the Euclidean distance and a dissimilarity index based on the coincidence similarity index are compared by using the k-neighbors supervised classification method respectively to features resulting from several types of transformations of one- and two-dimensional symmetric densities. Given two groups characterized by respective densities without or with overlap, different types of respective transformations are obtained and employed to quantitatively evaluate the performance of k-neighbors methodologies based on the Euclidean distance an coincidence similarity index. More specifically, the accuracy of classifying the intersection point between the densities of two adjacent groups is taken into account for the comparison. Several interesting results are described and discussed, including the enhanced potential of the dissimilarity index for classifying datasets with right skewed feature densities, as well as the identification that the sharpness of the comparison between data elements can be independent of the respective supervised classification performance.
\end{abstract}

\section{Introduction}\label{sec:introduction}

Pattern recognition (e.g.~\cite{duda2000pattern,theodoridis2006pattern,da2018shape}) basically consists of the endeavor of assigning categories to data elements. Remarkably, most human (and even animal) cognitive activities can be ultimately understood as specific instances of pattern recognition. For instance, as an individual moves in an environment, it becomes necessary to identify the surrounding objects, including a suitable place where to proceed with next step, as well as objects of possible interest (e.g.~an apple), as well as potential hazards (e.g.~sharp objects). The pattern recognition aspects underlying human activity are so substantial to the point that it becomes hard to identify respective human activities which cannot be understood as pattern recognition. Even the challenging ability of prediction, one of the main objectives underlying science, can be actually understood from that perspective. It is therefore hardly surprising that the area of pattern recognition has motivated such great interest and related works in science and technology.

Pattern recognition approaches, also known as data classification, can be subdivided into at least two major types: (i) \emph{supervised}, when a subset of the data with known categories is used for a training stage; and (ii) \emph{non-supervised}, when no previous knowledge is available about the data categories. The attention in the present work is focused on supervised pattern recognition. Needless to say, the latter type of pattern recognition can be substantially more challenging.

Typical approaches to pattern recognition involve first representing each data element to be categorized in terms of a respective set of \emph{measurements}, \emph{features}, \emph{properties}, or \emph{attributes}, term which are often used interchangeably. The choice of a specific set of features establishes a corresponding \emph{feature space} having each axis associated to each of the features. The issue of choosing a proper set of features respectively to a given specific pattern recognition problem constitutes one of the greatest challenges in this area, as accounted by various aspects. First, we have that having more measurements does not necessarily contribute to achieving more accurate recognition. That is particularly the case when the added features tend to be redundant to those already chosen, or when they are noisy or have little sensitivity.

Another important problem in feature selection is that there is a virtually infinite number of possible features that can be employed to characterize a given dataset. As a simple illustration, it suffices to realize that if $x$ is a possible feature, so are respectively transformed features $x^\alpha$, with $\alpha$ being a real value. This example is a particular case of an invertible \emph{feature transformation} (or mapping) of the general form:
\begin{align}
    y = f(x)
\end{align}

where $x$ is a primary feature and $y$ is a respectively derived new feature.

The primary feature can then be readily recovered from $y$ by using the inverse transformation (provided it exists), i.e.:
\begin{align}
    x = f^{-1}(y) 
\end{align}

which can be understood as a feature transformation on itself.

Other examples of feature transformations include but are not limited to, unit changes, translation, scaling, rigid body transformations, and taking logarithms.

Interestingly, the use of a primary feature or respectively derived features can have a major impact on the subsequently performed pattern recognition. Just as a simple example, representing a given property in millimeters will tend to increase its relative weight than it would have in case it would otherwise be measured in meters. In addition, as will be discussed in the present work, to the features magnitude, their respective density can also strongly impact, as will be discussed in the present work, subsequent classification approaches.

Another important type of features transformation concerns respective \emph{normalizations}, such as the \emph{standardization} (e.g.~\cite{JohnsonWichern,gewers2021principal}). This type of transformation can be employed in order to obtain new features that have similar ranges of variation, therefore addressing situations such as the above observed selection between millimeters and meters. 

Another particularly important issue related to features selection and transformation concerns the methodologies to be subsequently adopted for categorizing the data elements into respective categories. Henceforth, we assume that the data elements are supposed to be pairwise compared in terms of distance or similarity indices. In particular, we shall consider the supervised pattern recognition approaches known as \emph{Bayesian classification} and \emph{k-neighbors} (e.g.~\cite{duda2000pattern,theodoridis2006pattern,da2018shape}). The former involves comparing the values of the features densities weighted by the respective mass probabilities so that the category of the density leading to the largest density point for an element to be classified is taken as result.  For simplicity's sake, all situations considered in the present work are assumed to be equally likely, in the sense of having the same mass probabilities.   

The Bayesian methodology, which is extensible to any number of features and categories, is known to be optimal --- in the sense of allowing the smallest chance of data misclassification --- provided the densities are known to full accuracy. The latter method consists of, given a new data element $i$ to be categorized, identifying the $k$ trained data elements that are more closely related (e.g.~smaller distance or largest similarity) to the data element $i$. The category to be assigned to the new data element is then taken as corresponding to the category predominant among the $k$ nearest-neighbors. Interestingly, the $k-$neighbors approach tends to optimal performance, becoming equivalent to Bayesian decision, when an infinite number of training samples are considered.

\begin{figure}
  \centering
     \includegraphics[width=0.65 \textwidth]{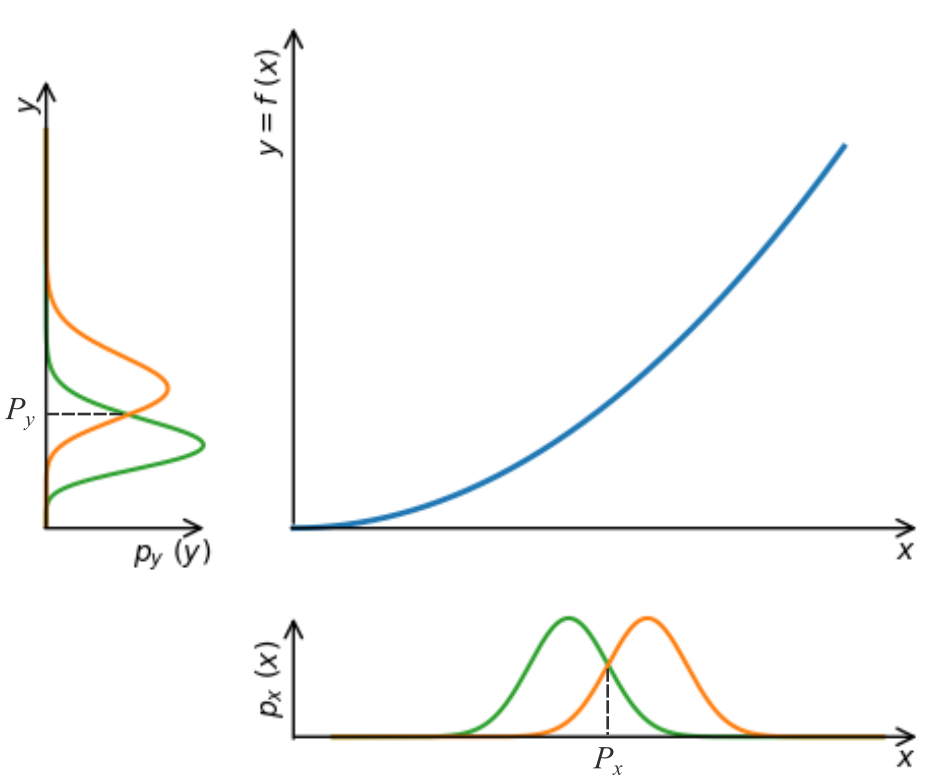}
 \caption{Two groups, shown in organge and green, are represented in terms of respective normal feature densities (symmetric around the average) on the variable $x$. The transformation of the measurement (feature) $x$ results in a new measurement $y$ characterized by skewed densities. The application of Bayesian decision considering  the feature $y$ requires the estimation of these relatively more complex densities.}\label{fig:transformation2}
\end{figure}

The adoption of specific types of features and transformations employed to represent the given set of data elements can have an important impact on the adopted classification method. As an illustration, consider the situation depicted in Figure~\ref{fig:transformation2}. Here, two equally likely categories of data elements characterized by their respective normal densities of a feature $x$ are transformed into a new feature $y$ by a generic function $f(x)$. The application of the Bayesian method indicates that the optimal decision regions are defined as being delimited by the point $P_x$ along the $x$ axis corresponding to the intersection of the two respective densities. However, when represented along the $y$ space, densities other than the normal need to be inferred in order to obtain the optimal decision point $P_y$. 

The present work aims at studying how the Bayesian decision and $k-$neighbors methods relying on Euclidean distance or multiset similarity indices perform respectively to varying types of features and transformations characterized by \emph{skewed densities} in one and two dimensions.  More specifically, we shall focus attention on \emph{right} (or positive) skewed densities.  This issue is of particular importance because the Euclidean distance metric often adopted in supervised classification methodologies is intrinsically symmetric and therefore not particularly suitable for dealing with skewed data.
In addition, as it will be discussed in Section~\ref{sec:skewed}, the density of features in real-world or abstract problems are often characterized by skewness.

The present work starts by discussing the important issue of skewed features densities, which is followed by a brief revision of the main basic concepts underlying the reported studies. Experiments involving the quantification of the effect of different types of skewed densities are then presented and discussed, followed by the identification of prospects for further related investigations.

\section{Right Skewed Feature Densities}\label{sec:skewed}

Interestingly, several features of real-world and abstract entities are intrinsically asymmetric, or skewed (e.g.~\cite{maruo2017,clauset2009,bonabeau1999,klaus2011,newman2005}). This important tendency has several origins, including but not being limited to the fact that features are often the result of non-linear transformations of otherwise uniform features (e.g.~the area of a square is equal to its squared size); the fact that several features need to be constrained along the non-negative (or non-positive) portion of the coordinate axes (e.g.~ages, length, width, etc.); as well as being implied by proportional features~\cite{benatti2024agglomerative} characterized (e.g.~same relative tolerance or dispersion of the length of ants and whales).  

Even though several features are intrinsically skewed, not so many related works have bee reported addressing this important issue.  Some of the most closely related approaches reported in the literature include~\cite{phua2004,hubert2008,gupta2008,hubert2009,maruo2017,godase2012}, but are by no means limited to these references.

\begin{figure}
  \centering
     \includegraphics[width=0.99 \textwidth]{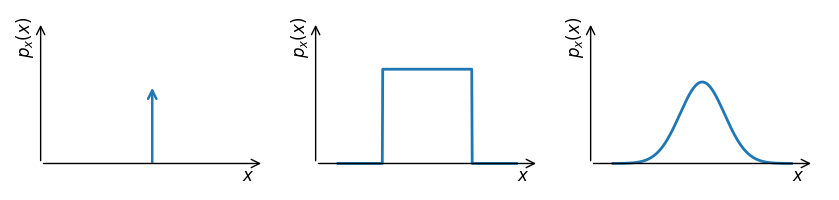} \\
     \hspace{.1cm} (a) \hspace{3.7cm} (b) \hspace{3.7cm} (c)
 \caption{Examples of symmetric feature densities on a random variable $x$: (a) constant (modeled as Dirac's delta); (b) uniform; and (c) normal. }\label{fig:densities}
\end{figure}

The features adopted for the characterization of datasets can be statistically modeled in terms of symmetric \emph{density functions} (e.g.~\cite{fisher1970,da2018shape}). Three of the main types of such densities, called \emph{constant}, \emph{uniform}, and \emph{normal}, are illustrated in Figure~\ref{fig:densities}(a), (b) and (c), respectively.

These densities are fully (bilaterally) symmetric respectively to their center of mass, or average, which is intrinsically compatible with the adoption of symmetric comparison operations, including the Euclidean distance.

\begin{figure}
  \centering
     \includegraphics[width=0.99 \textwidth]{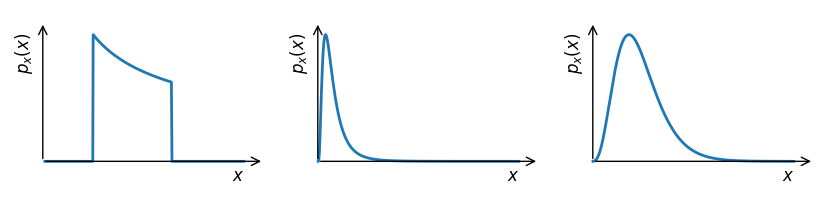}\\
     \hspace{.1cm} (a) \hspace{3.7cm} (b) \hspace{3.7cm} (c)
 \caption{Three examples of skewed feature densities.}\label{fig:skewness}
\end{figure}

However, there are several situations in which the data features densities present skewness, as illustrated in Figure~\ref{fig:skewness}.

The type of skewed densities $p(x)$ considered in the present work have the density values \emph{monotonically decreasing} with the value of the random variable $x$.

\begin{figure}
  \centering
     \includegraphics[width=0.45 \textwidth]{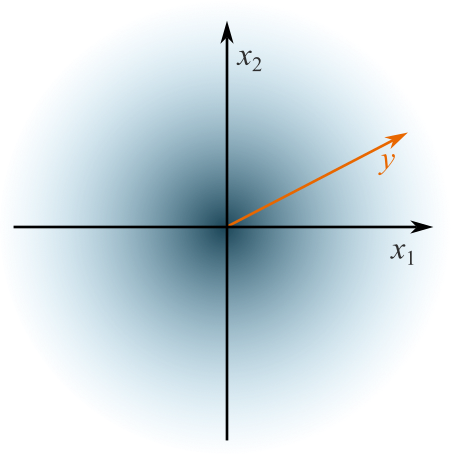}
 \caption{A possible model of two-dimensional dataset represented in terms of two features $x_1$ and $x_2$. The center of the cluster corresponds to the highest data density, which then decreases monotonically with the distance from that central point.}\label{fig:slice}
\end{figure}

There are many ways in which skewed feature densities can be obtained. First, we have the fact that a dataset is often characterized by the general type of varying density illustrated in Figure~\ref{fig:slice}.

This type of single-cluster model reflects the fact that the data elements tend to concentrate around the respective average, which therefore achieves its highest data density at the center of the respective feature space. The data density then tends to decrease monotonically with the distance from the average, leading to progressively less dense regions.  

\begin{figure}
  \centering
     \includegraphics[width=0.5 \textwidth]{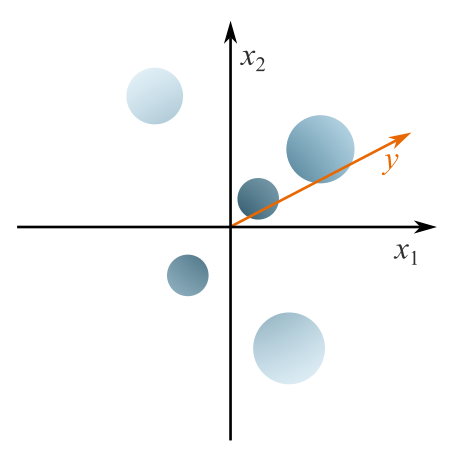}
 \caption{An example of clusters existing in a skewed feature space. Observe that the density of the clusters tends to decrease with their distance from the origin of the coordinate system $(x_1, x_2)$}\label{fig:clusters}
\end{figure}

Assuming that the skewness in the above example is implied intrinsically by the types of features $x_1$ and $x_2$, the eventual presence of clusters in this feature space will be similar to the situation illustrated in Figure~\ref{fig:clusters}, where 5 circular clusters can be identified with points at varying distances from the feature space origin. Importantly, those clusters closer to the original will therefore be characterized by higher data density, which will decrease monotonically with the distance from the origin. 

For simplicity's sake, we henceforth focus on the analysis of the densities along an \emph{one-dimensional domain} such as that illustrated by the orange slicing axis identified by the free variable $y$ in Figure~\ref{fig:slice}. The more general situation considering all the four quadrants can be readily taken into account by considering higher dimensional similarity operators such as that expressed in Equation~\ref{eq:coinc}.

The specific manner in which an one-dimensional data density decreases with the distance from the origin can vary substantially, provided it proceeds in monotonic manner. One interesting approach to specifying the data density variation consists of starting with a uniform feature density along an one-dimensional interval $[a,b]$ along a variable $x$  and then applying a transformation on this variable that leads to a specific monotonic density decrease (see also~\cite{ferreira2006}). The following such transformations henceforth adopted in this work are:
\begin{align}
   &y = f(x) = 2^x;  \label{eq:f1} \\
   &y = f(x) = x^2;  \label{eq:f2} \\
   &y = f(x) = x^3;  \label{eq:f3} \\
   &y = f(x) = e^{\alpha \, x};  \label{eq:f4} \\
   &y = f(x) = x  \label{eq:f5}  
\end{align}

Down concavity monotonic transformation functions (e.g.$f(x) =\sqrt{x}$) have not been considered in the present work, which focus attention on up concavity monotonic functions such those in Equations~\ref{eq:f1}--~\ref{eq:f4}.

Observe that the latter transformation above is not skewed, being considered here only for the sake of comparison.  All other transformations are skewed, also increasing monotonically.

\begin{figure}
  \centering
     \includegraphics[width=1 \textwidth]{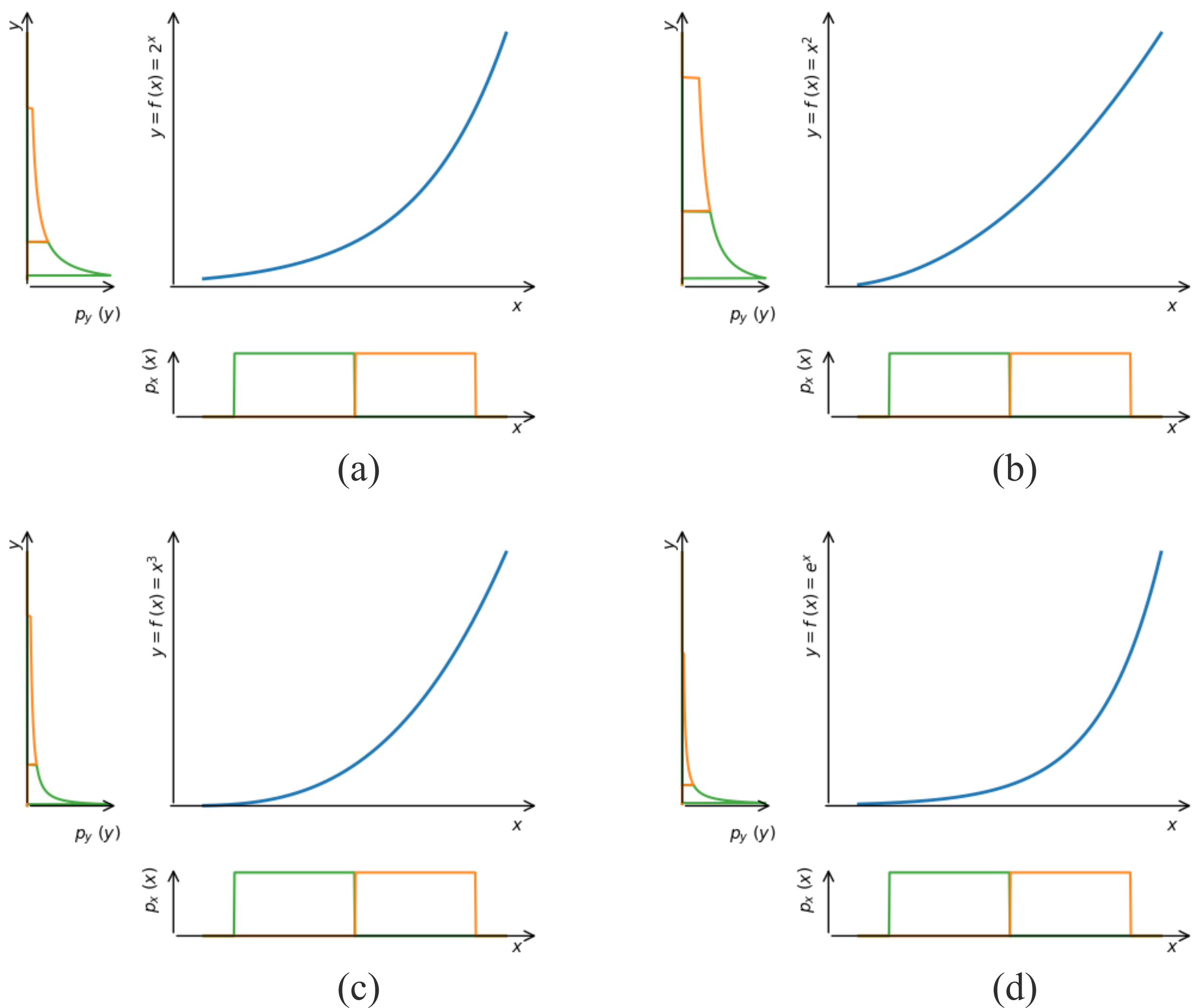}
 \caption{Illustrations of how the skewed feature densities considered in the present work have been obtained by transformation of two adjacent symmetric uniform original densities.}\label{fig:transformation_f}
\end{figure}

Figure~\ref{fig:transformation_f} illustrates skewed densities obtained by transforming a pair of adjacent uniform original densities by employing the transformations described in Equations~\ref{eq:f1}--\ref{eq:f4}.

The above described approach of obtaining skewed densities by transformations of symmetric original densities can be readily extended to obtaining skewed multivariate densities which can be expressed as the product of univariate densities, being therefore \emph{separable}.  In this case, the transformations are independently applied on the original univariate densities, and the results are then multiplied in order to obtain the multidimensional skewed densities.

\section{Basic Concepts}

This section presents some of the main basic concepts and methods employed in the present work, including the generation of skewed densities, the adopted supervised classification approaches, as well as the coincidence similarity index and some of its properties.

\subsection{Generating Skewed Densities}
Datasets with $N$ data elements, represented in terms of their $M$ respective individual \emph{features} $x_1, x_2, \ldots, x_M$, are henceforth assumed in the present work. Therefore, each data element $i$ can be expressed in terms of its respective \emph{feature vector}:
\begin{align}
  \vec{x}_i = \left[x_{i,1}, x_{i,2} \ldots, x_{i,M}  \right]
\end{align}

The \emph{Euclidean distance} between two feature vectors $\vec{x}_i$ and $\vec{x}_j$ can be expressed as:
\begin{align}
  E(\vec{x}_i, \vec{x}_j) = | \vec{x}_i - \vec{x}_j |= \sqrt{ \sum_{k=1}^M  \left(x_{i,k}-x_{j,k} \right)^2  }
\end{align}

Observe that, in principle, the Euclidean distance is not upper-bounded, i.e.~$0 \leq   E(\vec{x}_i, \vec{x}_j) < \infty$.

In principle, the adoption of the Euclidean distance as a quantification of the lack of relationship between data elements assumes that all features originally have the same physical unit, which is also the unit of the obtained distance results. In addition, the feature space needs to be a metric space itself, in the sense of each feature axis being orthogonal to all other axis. Generalized forms of distance adapted to affine spaces are also possible, but this requires the basis of these spaces to be known. Yet another important point to be taken into account concerns the fact whether the relationships between the features are to be considered in uniform or proportional manner~\cite{benatti2024agglomerative}.

The datasets may contain a total of $G$ groups labeled as $A$, $B$, $\ldots$. 
For simplicity's sake, it is henceforth assumed that $G=2$. In the case of univariate densities, each data element is represented in terms of a single feature $y$. The density function of the two groups $A$ and $B$ can be expressed as $p_A(y)$ and $p_B(y)$. Figure~\ref{fig:transformation_f}(b) illustrates a specific situation characterized by the presence of two adjacent groups represented by a feature $x$ following the uniform density being transformed into respective densities on the $y$ random variable by the transformation $y=f(x)=x^2$.

It can be shown that the new density $p_y(x)$obtained by transforming an initial density $p_x(x)$ on the random variable $x$ by a function $y=f(x)$ can be expressed as: 
\begin{align}
   p_y(y=f(x)) = \frac{1}{f'(x)} \, p_x(x),\label{eq:tr_x}
\end{align}
where $f'()$ is the first derivative of $f()$ respectively to $x$.

It is also possible to rewrite the above equation as:
\begin{align}
   p_y(y) = \frac{1}{f'(f^{-1}(y))} \, p_x(f^{-1}(y)),
\end{align}

\begin{figure}
  \centering
     \includegraphics[width=0.5 \textwidth]{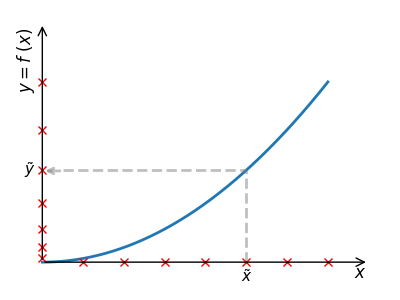}
 \caption{The mapping of a primary feature $x$ into a derived feature $y$ by using the particular transformation $y=f(x)=x^2$. Interestingly, this transformation has an important effect in changing the density of the elements along the $y$ space.}\label{fig:transformation}
\end{figure}

Equation~\ref{eq:tr_x} can be conceptually understood by considering the transformation of equally separated points $x$, as illustrated in Figure~\ref{fig:transformation}. Each of these points is mapped into respective new values $y$ by the transformation $y=f(x)=x^2$. The lower the derivative of the function $y = f(x)$ at a given point $\tilde{x}$, the higher the density around the respective image $\tilde{y}$.

The \emph{Bayesian decision} theory of supervised classification indicates that the probability of misclassifications of a new data element with value $x_n$ can be minimized provided the category associated to the density, between $p_A(y)$ and $p_B(y)$, is taken as the result. Figure~\ref{fig:transformation2} illustrates a situation in which the respective decision regions are determined from the value $P_x$ where the two group densities intersect one another. Interestingly, it can be shown that optimal classification can be achieved provided the densities are known in a fully accurate manner.

\subsection{Supervised Classification}
The supervised methodology known as \emph{k-neighbors} (e.g.~\cite{duda2000pattern,theodoridis2006pattern,da2018shape}) can be applied without the need to have the parametric forms of the groups densities. More specifically, a set of $T$ training points of both groups $A$ and $B$ are provided associated to the respective categories. Though these points are sampled from respective densities, the latter may not be known. Then given a new data element with $x_n$, the decision criterion consists simply of identifying the $k$ most closely related trained points (neighbors) and taking as resulting category that corresponding to the larger number of identified neighbors. Both the Euclidean distance and coincidence similarity index can be used in order to identify the set of $k$ most related (similar) neighbors.
Interestingly, the performance of the $k-$neighbors methodology tends to the optimal performance when the number of training points becomes larger and larger.

\subsection{The Coincidence Similarity Index}

The coincidence similarity index has been described (e.g.~\cite{da2021further,costa2022simil,da2022coincidence,da2024integrating,costa2023mneurons}) as a generalization of the traditional Jaccard index (e.g.~\cite{Jac:wiki,da2021further}) allowing the consideration of the relative interiority between the compared sets to be taken into account, and also for comparing two real-valued vectors, understood in the sense of multiset theory (e.g.~\cite{da2021multisets}).

Given two real-valued vectors $\vec{u}$ and $\vec{v}$, at least one of them being non-zero, with $M$ coordinates each, their coincidence similarity can be estimated as follows.  First, each of the two vectors are represented in terms of respective np-sets~\cite{da2024integrating}, i.e.:
\begin{align}
   &\vec{u}   \longleftrightarrow \left\{  [1, m^p_{u,1}, m^n_{u,1}];  [2, m^p_{u,2}, m^n_{u,2}]; \ldots;  [M, m^p_{u,M}, m^n_{u,M}]  \right\}   \\
   &\vec{v}   \longleftrightarrow \left\{  [1, m^p_{v,1}, m^n_{v,1}];  [2, m^p_{v,2}, m^n_{v,2}]; \ldots;  [M, m^p_{v,M}, m^n_{v,M}]  \right\}   
\end{align}

where:
\begin{align}
  &m^p_{u,k} = \max \left(u_k, 0 \right) \\
  &m^n_{u,k} = \min \left(u_k, 0 \right) \\
  &m^p_{v,k} = \max \left(v_k, 0 \right) \\
  &m^n_{v,k} = \min \left(v_k, 0 \right) 
\end{align}

Now, the coincidence similarity index between the two vectors can be expressed as:
\begin{align}
   \mathcal{C}(\vec{u},\vec{v}) =& \, \mathcal{C}(\vec{v},\vec{u}) =
   \left[ \frac{|\vec{u} \cap \vec{v}|}{|\vec{u} \cup \vec{v}|}\right]^D \ \frac{|\vec{u} \cap \vec{v}|}{\min( |\vec{u}|, |\vec{v}|)} =
   \nonumber \\
   & = \left[ 
   \frac{ \sum_{k=1}^{M} \left[ \min \left( m^p_{u,k}, m^p_{v,k} \right)  + \min \left( |m^n_{u,k}|, |m^n_{v,k}| \right) \right]  }
   { \sum_{k=1}^M \left[ \max \left( m^p_{u,k}, u^p_{v,k} \right)  + \max \left( |m^n_{u,k}|, |m^n_{v,k}| \right) \right] }  \right]^D
   \times  \nonumber \\
   & \hspace{1cm}  \times  \left[ \frac{\sum_{k=1}^{M} \left[ \min \left( m^p_{u,k}, m^p_{v,k} \right)  + \min \left( |m^n_{u,k}|, |m^n_{v,k}| \right) \right] }
   {\min \left( \sum_{k=1}^{M} \left[ m^p_{u,k} +|m^n_{u,k}| \right], \sum_{k=1}^{M} \left[ m^p_{v,k} +|m^n_{v,k}| \right]  \right)  }  \right]^E
   \label{eq:coinc}
\end{align}

where $D$ and $E$ are non-negative real value exponents controlling how sharp the implemented comparison is.    The larger the values of $D$ and $E$, the sharer and more strict the comparisons are, with parameter $E$ controlling the width of the respective comparison region.  It is henceforth adopted that $E=1$. It can be shown that $0 \leq \mathcal{C}(\vec{u},\vec{v}) \leq 1$.

For simplicity's sake, the present work henceforth assumes all features to be non-negative, in which case the above expression can be simplified as:
\begin{align}
   &\mathcal{C}(\vec{u},\vec{v}) = \mathcal{C}(\vec{v},\vec{u}) = \nonumber \\
   & = \left[ \frac{\sum_{k=1}^{M} \min \left( u_{k}, v_k \right) }
   {\sum_{k=1}^{M} \max \left( u_{k}, v_k \right)}  \right]^D  
   \,
   \left[ \frac{ \sum_{k=1}^{M} \min \left( u_{k}, v_k \right) }
   {\min \left( \sum_{k=1}^{M} |u_k|, \sum_{k=1}^{M} |v_k|\right)  }  \right]^E \label{eq:coincp}
\end{align}

Unlike the Euclidean distance, the adoption of the coincidence similarity does not require the original features to be originally represented in the same units.  Neither is that space required to be metric or orthogonal. This property follows from the fact that the coincidence similarity is based on \emph{multiset operations}, more specifically \emph{np-sets}, which are analogous to sets.  
The similarity comparison is performed in relative terms, so that a non-dimensional similarity index is obtained.  More specifically, we have that:
\begin{align}
      &\mathcal{C}(\gamma \vec{u}, \gamma \vec{v}) = \nonumber \\
   & = \left[ \frac{\sum_{k=1}^{M} \gamma \min \left( u_{k}, v_k \right) }
   {\sum_{k=1}^{M} \gamma \max \left( u_{k}, v_k \right)}  \right]^D  
   \,
   \left[ \frac{ \sum_{k=1}^{M} \gamma \min \left( u_{k}, v_k \right) }
   {\min \left( \gamma \sum_{k=1}^{M} |u_k|, \gamma \sum_{k=1}^{M} |v_k|\right)  }  \right]^E = 
   \nonumber \\ 
   & = \mathcal{C}(\vec{u}, \vec{v})  
\end{align}

where $\gamma$ is a positive real value.

This type of comparison is intrinsically compatible with proportional features~\cite{benatti2024agglomerative}, or  when two quantities need to be compared in relative terms so as to be invariant with the scaling of the resolution or dispersion of the data by a given constant $\gamma$.

For the sake of compatibility in the comparisons with the Euclidean distance to be performed in this work, a \emph{dissimilarity index} is henceforth obtained from the coincidence similarity (e.g.~\cite{costa2022simil}) as:
\begin{align}
  \Delta(\vec{u},\vec{v}) = \Delta(\vec{v},\vec{u}) = 1 - \mathcal{C}(\vec{v},\vec{u})
\end{align}

As with the coincidence similarity index, we also have that $0 \leq \Delta(\vec{v},\vec{u}) \leq 1$

\begin{figure}
  \centering
     \includegraphics[width=0.9 \textwidth]{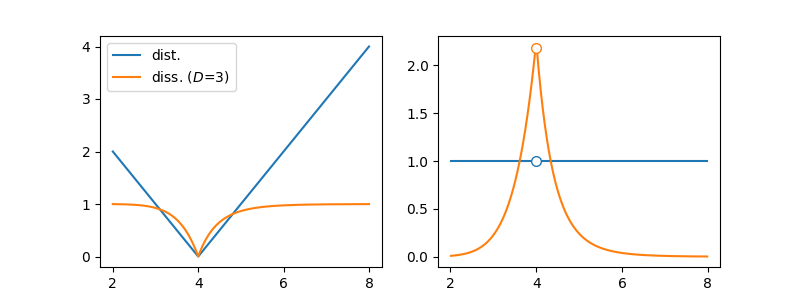}\\
     \hspace{.2cm} (a) \hspace{4cm} (b)
  \caption{The profiles of the Euclidean distance (dist.) and dissimilarity index (diss.) for comparisons with the reference value $4$ are shown in (a).  The respective sensitivity, expressed in terms of the derivatives of the profiles in (a) are presented in (b).  The circles on the functions at $y=4$ indicate that the respective derivatives cannot be obtained at that point.}\label{fig:sensitivity}
\end{figure}

A particularly important property of an operator for comparing two vectors, as is the case of the Euclidean distance and dissimilarity index, concerns how \emph{sharp} these operators are around their respective peak, a property that defines how strict the performed comparison is. Figure~\ref{fig:sensitivity}(a) illustrates the profiles of the Euclidean distance and the adopted dissimilarity index, based on the coincidence similarity for $D=3$ respectively to comparing real values with the reference value $4$.

The sharpness of each profile can be objectively quantified in terms of the absolute value of the first derivative  --- henceforth called \emph{sensitivity}~\cite{costa2023mneurons} --- of the obtained profiles, as illustrated in Figure~\ref{fig:sensitivity}(b). It can be readily verified that the dissimilarity index has substantially enhanced sensitivity around the comparison value $4$. The sensitivity of the coincidence dissimilarity index can be controlled by the parameter $D$.

It follows from the obtained sensitivity values that, as expected (e.g.~\cite{costa2022simil,costa2023mneurons}), the dissimilarity index is substantially more sensitive around the comparison peak than than the Euclidean distance.

\section{Experiments and Discussion}

In this section, the potential of the Euclidean distance and the dissimilarity index for categorizing data elements is experimentally studied respectively to several types of features densities obtained by transforming a reference features $x$ with uniform density involving two adjacent groups $A$ and $B$ intersecting at $x=P_x$. The groups are sampled from the original uniform density $p(x)$ with $N=N_A+N_B$ points. The considered densities correspond to those obtained by the transformations of uniform densities as specified by Equations~\ref{eq:f1}--\ref{eq:f5}: power functions, quadratic, cubic, exponential, and linear.  Experiments have been performed also in the case of two-variate densities.

The experiment consists in quantifying, by using the Euclidean distance and the dissimilarity index, the total numbers $n_A$ and $n_B$ of neighbors belonging to the two original groups that resulting when considering the $k$ neighbors most related to the value of $P_y=f(P_x)$. Because $P_y$ corresponds to the intersection of the densities $p_A(y)$ and $p_B(y)$, it should ideally follows that $n_A= n_B$. Thus, the ability of the Euclidean distance and dissimilarity indices for identifying this separation point $P_x$ can be quantified in terms of the following \emph{accuracy index}:
\begin{align}\label{eq:acc}
   \beta(n_A,n_B) = \frac{ \min \left(n_A,n_B \right) } { \max \left(n_A,n_B \right)},
\end{align}
which turns out to correspond to the Jaccard similarity index between the averages $<n_A>$ and $<n_b>$ over $R=1,000$ realizations, so that $0 \leq \beta(<n_A>,<n_B>) \leq 1$ (assuming that $n_A > 0$ and $n_B > 0$) and $\beta(n_A,n_B)=\beta(<n_B>,<n_A>)$. Ideally, $\beta(<n_a>,<n_B>)$ should result equal to 1, confirming that point $P_y$ corresponds to the intersection between the two densities $p_A(y)$ and $p_B(y)$.

It is interesting to observe that, given that $n_A$ and $n_B$ correspond to positive integer values, it follows that $\beta(n_A,n_B)$ necessarily takes discrete (quantized) values.

A particularly interesting point concerns the fact that the accuracy experiments specifically performed in this work do not depend on the choice of the parameter $D>0$. That is because the order and monotonicity of the closest neighbors obtained by using distinct values of $D$ in Equation~\ref{eq:coinc} do not change with that parameter. Or, in other words, we have that:
\begin{align}
    \mathcal{C}(\vec{u},\vec{v}) \geq \mathcal{C}(\vec{r},\vec{s}) \Longleftrightarrow
    \left[ \mathcal{C}(\vec{u},\vec{v}) \right] ^{\alpha} \geq \left[ \mathcal{C}(\vec{r},\vec{s}) \right]^{\alpha} 
\end{align}

where $\alpha \in \mathcal{R}$, $\alpha > 0$.

This result indicates that the sharpness implemented by the dissimilarity index can be independent of the respective classification accuracy. It follows that the dissimilarity index paves the way to obtaining an approach that can both implement sharp (strict) comparisons at the same time of providing enhanced results in the case of datasets characterized by having skewed feature densities.

\begin{figure}
  \centering
     \includegraphics[width=0.99 \textwidth]{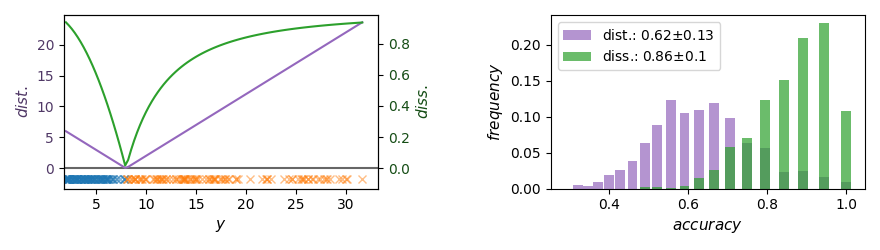}\\
     \hspace{.8cm} (a) \hspace{6.6cm} (b)

     \includegraphics[width=0.99 \textwidth]{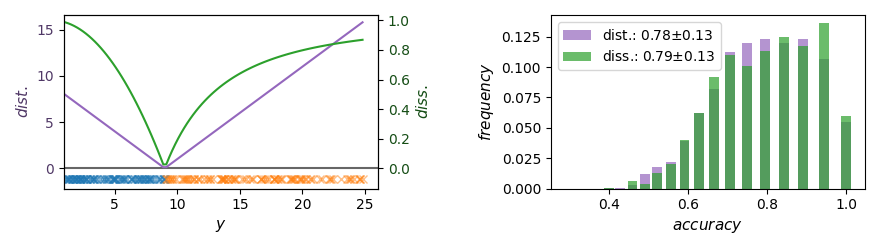}\\
     \hspace{.8cm} (c) \hspace{6.6cm} (d)
     \includegraphics[width=0.99 \textwidth]{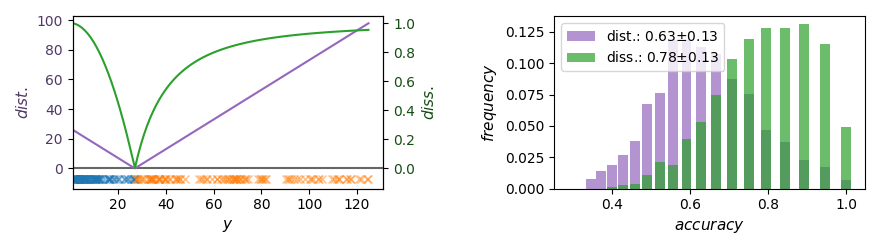}\\
     \hspace{.8cm} (e) \hspace{6.6cm} (f)
     \includegraphics[width=0.99 \textwidth]{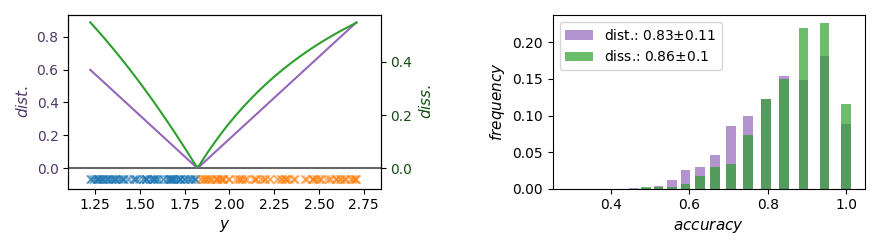}\\
     \hspace{.8cm} (g) \hspace{6.6cm} (h)
 \caption{Classification accuracy around the intersection point obtained by the $k-$neighbors ($k=70$) methodology employing the Euclidean distance (dist.) and dissimilarity index $diss.$ assuming $D=3$ for feature densities obtained from several types of transformations of uniform initial densities: $f(x)=2^x$ (a,b); $f(x)=x^2$ (c,d); $f(x)=x^3$ (e,f); $f(x)=\exp(0.2x)$ (g,h). The densities shown at the right-hand side of this figure were obtained respectively to $R=1,000$ experiments. The similarity operation allowed better performance in all considered situations.}\label{fig:methods}
\end{figure}

Figure~\ref{fig:methods} depicts the main result from $R=1,000$ realizations of the accuracy experiments respective to the transformations in Equations~\ref{eq:f1}--\ref{eq:f4}. The experiments were performed for $k=70$ and $N_A=N_B=100$. Shown at the left-hand side of the figure are examples of data element densities, together with the respective distance and dissimilarity curves centered at $P_y$. The obtained densities of the accuracy index $\beta$ are depicted at the right-hand side of the figure.

It can be readily verified that the dissimilarity approach yielded better accuracy in all the situations involving skewed densities considered in Figure~\ref{fig:methods}. This interesting result follows from the fact that the dissimilarity approach is intrinsically more compatible with the features skewness than the Euclidean distance. That is because the coincidence similarity index implements proportional comparisons, while the Euclidean distance is based on uniform comparisons~\cite{benatti2024agglomerative}.

\begin{figure}
  \centering
     \includegraphics[width=0.99 \textwidth]{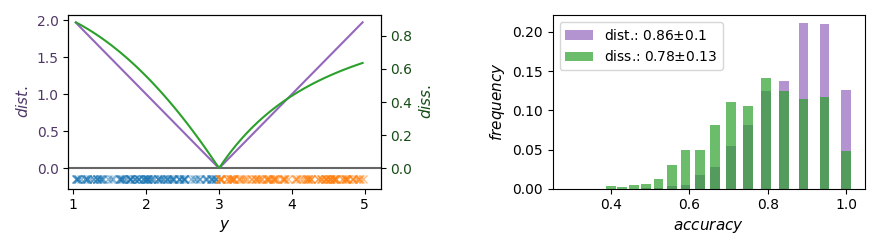}\\
     \hspace{.8cm} (a) \hspace{6.6cm} (b)
 \caption{Classification performance around the decision point in the case of two categories having uniform feature densities, quantified in terms of the accuracy index $\beta$, obtained for the Euclidean distance (dist.) and dissimilarity (diss.) comparison operations. The considered data samples are illustrated in (a) jointly with the respective distance and dissimilarity curves, while the histograms of the obtained accuracy index is shown in (b). Though the former operation yields better performance, as expected as a consequence of density symmetry, the dissimilarity approach still leads to a not substantially smaller value.}\label{fig:method_linear}
\end{figure}

For comparison purposes, accuracy results obtained respectively to the uniform densities of the initial variable $x$ are shown in Figure~\ref{fig:method_linear}, in which specific case the Euclidean distance approach led to better performance, though not by a substantial margin.

\begin{figure}
  \centering
     \includegraphics[width=0.495 \textwidth]{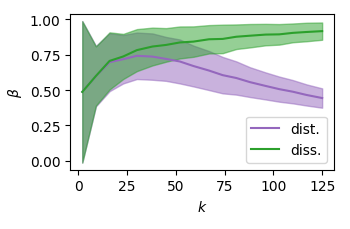}
     \includegraphics[width=0.495 \textwidth]{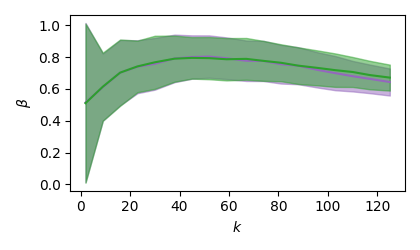} \\
     \hspace{.9cm} (a) \hspace{6cm} (b)
     \includegraphics[width=0.495 \textwidth]{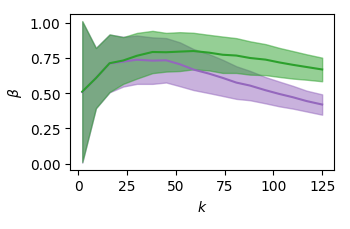}
     \includegraphics[width=0.495 \textwidth]{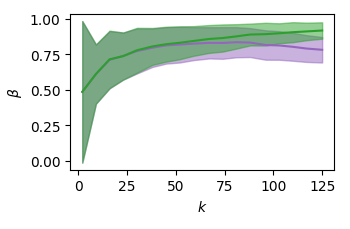}\\
     \hspace{.9cm} (c) \hspace{6cm} (d)
     \includegraphics[width=0.495 \textwidth]{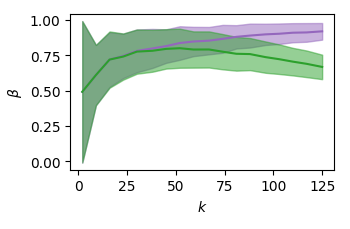}\\
     \hspace{.9cm} (e)
 \caption{The average $\pm$ values of the accuracy index $\beta$ in terms of the number of neighbors $k$ obtained for the experiments considered in this work respectively to supervised identification with $N_A=N_B=100$, by $k-$neighbors considering Euclidean distance and dissimilarity, of the intersection point between densities obtained from Equations~\ref{eq:f1}--\ref{eq:f5}: power (a), quadratic (b), cubic (c), exponential (d); and linear (e). The dissimilarity approach resulted improved or similar respectively to the Euclidean distance approach in all considered situations --- i.e.~ (1--d) -- involving skewed feature densities. At the same time, the Euclidean distance-based $k-$neighbors methodology has enhanced potential for classification involving symmetric densities (e).  It is also interesting to observe that the average curves tend to present a peak at intermediate values of $k$.
 }\label{fig:plot}
\end{figure}

Figure~\ref{fig:plot} presents the average $\pm$ standard deviation of the accuracy index $\beta$, in terms of several numbers of neighbors $k$, obtained for the Euclidean distance and dissimilarity index respectively to each of the considered feature transformations considering $R = 1,000$ and $N_A = N_B = 100$.

The dispersion of the skewed curves can be verified to present a tendency to decrease with the value of $k$. In addition, as verified for $k=70$ (see Fig.~\ref{fig:methods}) and respective discussion above), the dissimilarity index approach again yielded enhanced or similar classification accuracy in all considered cases involving skewed feature densities, particularly in cases of power transformation (a), cubic (c), and exponential (d). As before, the symmetric density (e) was better addressed by using the Euclidean distance methodology. 

Another interesting result that can be appreciated from Figure~\ref{fig:plot} concerns the fact that several of the curves shown present a peak at an intermediate value of $k$, which implies that the best accuracy depends not only on the type of comparison (i.e.~distance or dissimilarity) but also on the choice of the number of neighbors $k$ to be adopted.

One of the exceptions to presenting a peak of accuracy concerns the situation shown in Figure~\ref{fig:plot}(a), corresponding to skewness induced by the power transformation function $y = 2^x$. In this case, the accuracy allowed by the dissimilarity index increased monotonically. This is also the situation leading to the larger differences between the dissimilarity and distance curves. That is so because the power transformation of the original uniform densities leads to proportional densities, which is precisely the way in which the coincidence similarity performs its relative comparisons.

\begin{figure}
  \centering
     \includegraphics[width=0.495 \textwidth]{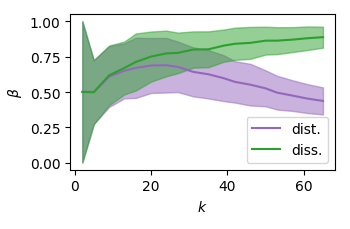}
     \includegraphics[width=0.495 \textwidth]{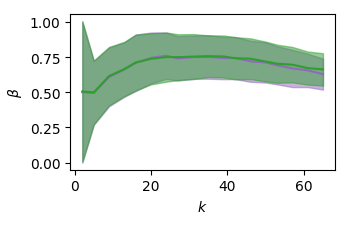} \\
     \hspace{.9cm} (a) \hspace{6cm} (b)
     \includegraphics[width=0.495 \textwidth]{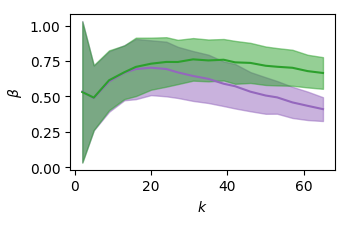}
     \includegraphics[width=0.495 \textwidth]{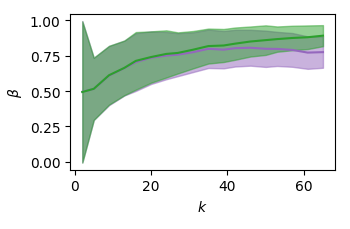}\\
     \hspace{.9cm} (c) \hspace{6cm} (d)
     \includegraphics[width=0.495 \textwidth]{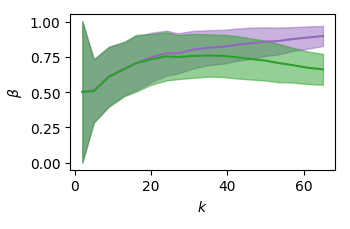}\\
     \hspace{.9cm} (e)
 \caption{Average $\pm$ standard deviation of the classification accuracy $\beta$ in terms of $k$ for each of the considered five types of densities while taking $N_A=N_B=50$: power transformation (a), quadratic (b), cubic (c), exponential (d), and linear (e). The obtained results are closely similar to those shown in Fig.~\ref{fig:plot}, with the dissimilarity approach leading to enhanced or similar results for all skewed densities (a--d).
 }\label{fig:plot_50}
\end{figure}

In order to consider the effect of taking another number of samples from each group, Figure~\ref{fig:plot_50} presents results of the above described experiment for $N_A=N_B=50$. The results are closely similar to those obtained for the previous experiment ($N_A=N_B=100$) shown in Figure~\ref{fig:plot}.

\begin{figure}
  \centering
     \includegraphics[width=0.495 \textwidth]{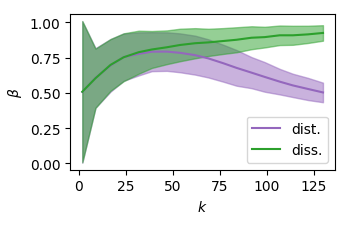}
     \includegraphics[width=0.495 \textwidth]{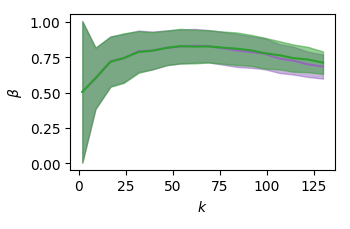} \\
     \hspace{.9cm} (a) \hspace{6cm} (b)
     \includegraphics[width=0.495 \textwidth]{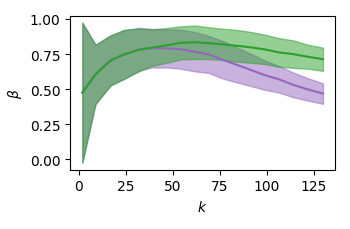}
     \includegraphics[width=0.495 \textwidth]{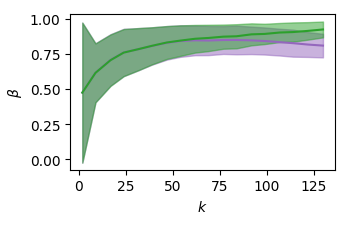}\\
     \hspace{.9cm} (c) \hspace{6cm} (d)
     \includegraphics[width=0.495 \textwidth]{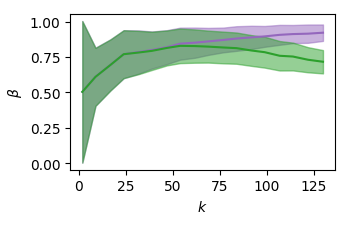}\\
     \hspace{.9cm} (e)
 \caption{Average $\pm$ standard deviation of the classification accuracy $\beta$ in terms of $k$ for each of the considered five types of densities obtained from transformations of two normal densities while considering $N_A=N_B=100$: power transformation (a), quadratic (b), cubic (c), exponential (d), and linear (e).  The obtained results are closely similar to those shown in Fig.~\ref{fig:plot}.
 }\label{fig:normal_plot}
\end{figure}

The accuracy experiment has also been performed considering skewed densities obtained from two \emph{overlapping} groups specified by transformation of respective normal densities. More specifically, the two original normal densities have respective averages $2.8$ and $4.0$ and standard deviations $0.333$. The resulting curves, obtained with $R=1000$ and $N_A = N_B = 100$ are presented in Figure~\ref{fig:normal_plot}.

It can be verified that these results are similar to those obtained in the two previous experiments, with the difference that the dissimilarity index tended to have its accuracy improved further.

To complement the three univariate experiments described above, an additional situation is now addressed involving a pair of two-dimensional skewed densities which can be expressed as the product of two univariate densities (being therefore a \emph{separable} scalar field), i.e.:
\begin{align}\label{eq:2Ddensity}
    p_{y_1,y_2} \left(y_1, y_2 \right) = p_{y_1}(y_1) \ p_{y_2}(y_2) = \nonumber \\
    = f(p_{x_1}(x_1)) \ f(p_{x_2}(x_2))
\end{align}

\begin{figure}
  \centering
     \includegraphics[width=0.49 \textwidth]{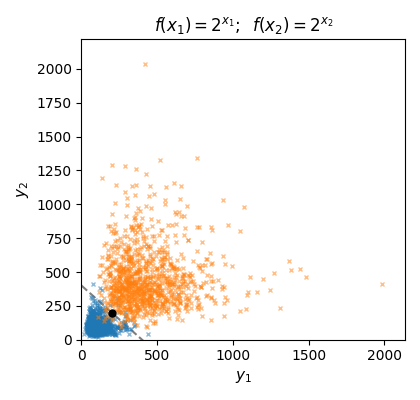}
     \includegraphics[width=0.49 \textwidth]{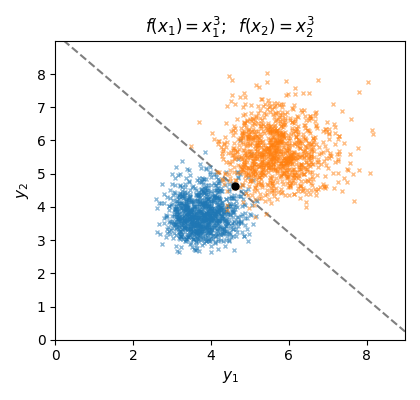} \\
     \hspace{1.1cm}(a) \hspace{6.1cm} (b)
 \caption{Illustrations of two-dimensional distributions of points obtained from skewed densities corresponding to two groups presenting some overlap, considering $N_A=N_B=1,000$. The transformations $f(x)=2^x$, depicted in (a), and $f(x)=x^3$, shown in (b), have been adopted for both independent variables $x_1$ and $x_2$. The central dot represents the reference value for the comparisons implemented by the Euclidean distance and coincidence similarity index. 
 The dashed line indicates the transformation by $f(x)$ of the intersection between the original normal densities of the form $p_{x_1,x_2}(x_1, x_2) = p_{x_1}(x_1) \ p_{x_2}(x_2)$.}\label{fig:2D_cluster}
\end{figure}

More specifically, we start with a pair (corresponding to groups $A$ and $B$) of symmetric normal densities on independent random variables $x_1$ and $x_2$ and apply the same random variable transformation over each of the variables. Figure~\ref{fig:2D_cluster} illustrates one such case corresponding to the transformations $f(x_1)=2^{x_1}$ and $f(x_2)=2^{x_2}$, shown in (a), and $f(x_1)=x_1^3$ and $f(x_2)=x_2^3$, presented in (b), of two univariate normal densities with averages $\left< x_1 \right> = 7 $ and $\left< x_2 \right> = 9$ and identical standard deviation $\sigma = 1$.

\begin{figure}
  \centering
     \includegraphics[width=0.495 \textwidth]{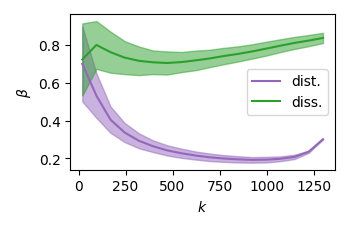}
     \includegraphics[width=0.495 \textwidth]{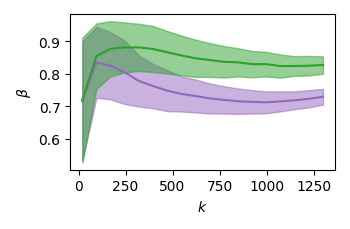} \\
     \hspace{.9cm} (a) \hspace{6cm} (b)
     \includegraphics[width=0.495 \textwidth]{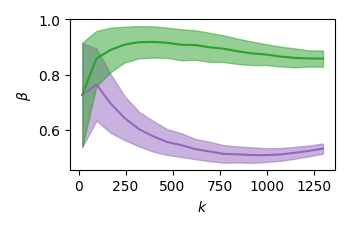}
     \includegraphics[width=0.495 \textwidth]{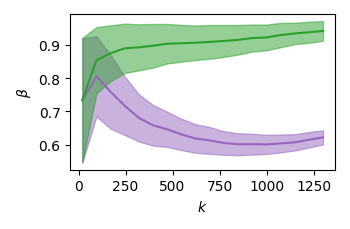}\\
     \hspace{.9cm} (c) \hspace{6cm} (d)
     \includegraphics[width=0.495 \textwidth]{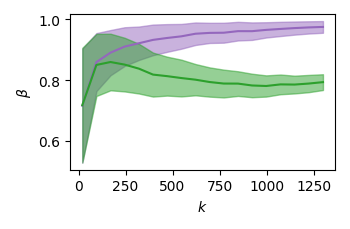}\\
     \hspace{.9cm} (e)
 \caption{Average $\pm$ standard deviation of the classification accuracy $\beta$ in terms of $k$ for each of the considered five types of densities obtained from transformations of a pair of two-dimensional normal densities of the type indicated in Eq.~\ref{eq:2Ddensity}, while considering $N_A=N_B=1,000$ obtained by the following transformations: power transformation (a), quadratic (b), cubic (c), exponential (d), and linear(e). The coincidence similarity resulted in substantially larger accuracy values for all cases involving skewed densities.
 }\label{fig:2D_normal_plot}
\end{figure}

The reference point for obtaining the Euclidean distance and coincidence similarity index accuracy comparisons corresponds to the transformation of the mid-point between the original normal densities. The experiment proceeded in similar manner as with the univariate cases, but now the Euclidean distance and coincidence similarity index are applied to two-dimensional arguments ($M=2$), corresponding to the coordinates $y_1$ and $y_2$ of each point and reference point. Equation~\ref{eq:coinc} is therefore employed for calculating the coincidence similarity indices. The obtained results are presented in Figure~\ref{fig:2D_normal_plot}.

\begin{figure}
  \centering
     \includegraphics[width=0.495 \textwidth]{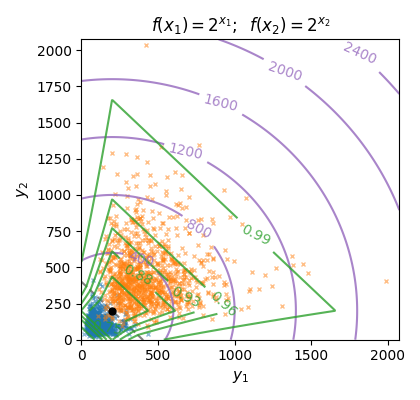}
     \includegraphics[width=0.495 \textwidth]{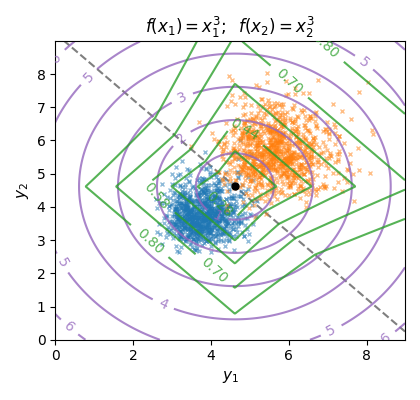} \\
     \hspace{.9cm} (a) \hspace{6cm} (b)
 \caption{Level sets of the Euclidean distance (violet) and dissimilarity index (green) superimposed on the distribution of points from Fig.~\ref{fig:2D_cluster}. This example shows how the proportional nature of the dissimilarity operator allows a more effective adaptation to the considered distribution of points.
 It is interesting to observe that diamond-shaped level sets would have been obtained in case $E=0$.
 }  \label{fig:2D_levels}
\end{figure}

Interestingly, the $k-$neighbor approach based on the coincidence similarity index allowed substantially enhanced accuracy. This results follows from the fact that, by implementing a proportional comparison, the dissimilarity index is intrinsically more adequate for coping with several types of skewed densities, as illustrated in Figure~\ref{fig:2D_levels} respectively to the distribution of points shown in Figure~\ref{fig:2D_cluster}, assuming $D=3$ and $E=1$.

The results obtained and discussed in this section suggest that the relative advantage of the coincidence similarity index in the case of skewed densities increases with the dimension of the data elements.

\section{Concluding Remarks}

One of the many challenges implied by pattern recognition consists in how to choose the features adopted for representing the data elements of interest. Though many approaches to pattern recognition intrinsically assume the features to be uniformly or normally distributed in respective feature spaces, other types of feature densities can often appear in practice. 

In the present work, we aimed at investigating how skewed densities can impact on supervised pattern recognition. More specifically, the $k-$neighbors methodology has been considered by adopting the two following alternative quantifications of data interrelationships: (a) Euclidean distance, and (b) a dissimilarity index based on the coincidence similarity index. In the latter case, the $k-$neighbors correspond to those data elements that are more similar to the reference point (intersection between the two group densities).

The considered experiments consisted of quantifying, in terms of a respective index $\beta$, the accuracy of $k-$neighbors supervised classification of the decision point separating the two groups, sampled with $N_A, N_B$ sample points from densities obtained by transforming uniform densities by using several types of functions. The accuracy of the classification has been obtained by taking into account the multiset Jaccard similarity index between the number $k$ of the neighbors closest to the transformed classification point which belonged to each of the two groups.

Several interesting results have been described and discussed. First, it has been verified that the dissimilarity approach allowed improved or similar accuracy in all considered cases involving skewed feature densities. Even in the specific case of a linear transformation involving univariate densities, the dissimilarity index allowed performance not substantially smaller than that obtained by the Euclidean distance.

Another interesting result concerns the fact that the sharpness (sensitivity) of the comparison implemented by the respective operations is not necessarily related to the accuracy of the respective supervised classification: while the parameter $D$ controls how strict the dissimilarity comparisons are, the obtained classification does not depend of that parameter. Interestingly, in this respect the described dissimilarity index, based on the coincidence similarity, allowed high sensitivity to be achieved allied with enhanced classification accuracy in several of the considered cases involving skewed densities. At the same time, dataset underlain by more uniform feature densities can be more effectively approached by using the Euclidean distance. These results suggest that preliminary quantifications of the density skewness, e.g.~in terms of higher order statistical moments, can provide a preliminary indication that can be taken as a subside for choosing between distance and dissimilarity comparison operations.

Of particular interest, we have that, as suggested by the performed experiments, the enhanced accuracy of the coincidence similarity applied to skewed data sets seems to increase with the dimensionality of the data elements. A similar effect occurs respectively to the enhanced performance of the Euclidean distance in the case of symmetric densities.

Though the balance of $k-$neighbors, reflected in the index $\beta$, has been described in the present work as a means of quantifying the accuracy potential of the two compared methodologies, it is interesting to observe that this approach can lead to a respective supervised classification method on itself: instead of considering the neighbors of a point for its classification (the typical $k-$neighbors approach), the point yielding the most balanced separation of the groups is used to define a respective decision reference.

It is important to keep in mind that the results and conclusions reported in this work are specific to the adopted densities, transformations, number of samples, intervals, as well as parameter configurations. It would also be interesting to perform complementary accuracy analysis, including other points in addition to that corresponding to the intersection between the densities, as well as consider classification methods other that the $k-$neighbors methodology. Additional further research possibilities motivated by the developments described in this work include but are not limited to considering feature spaces with dimension larger than two, other types of transformations, more than two groups, as well as unsupervised classification decision.

\section*{Acknowledgments}
A. Benatti are grateful to MCTI PPI-SOFTEX (TIC 13 DOU 01245.010\\222/2022-44), FAPESP (grant 2022/15304-4), and CNPq. Luciano da F. Costa thanks CNPq (grant no.~307085/2018-0) and FAPESP (grant 2022/15304-4).

\bibliography{ref}
\bibliographystyle{unsrt}

\end{document}